\documentclass[runningheads]{llncs}
\usepackage{graphicx}
\usepackage{graphics}
\usepackage{wrapfig}
\usepackage{float}
\usepackage{textcomp}
\begin{document}
\title{Non-locally Encoder-Decoder Convolutional Network for Whole Brain QSM Inversion}
%
%

\author{{Juan Liu\inst{1,2}} \and
Kevin M. Koch\inst{1,2,3}}
\authorrunning{J. Liu, K.M. Koch}
%
\institute{Center for Imaging Research, Medical College of Wisconsin, Milwaukee, WI, USA \and
Biomedical Engineering, Marquette and Medical College of Wisconsin, Milwaukee, WI, USA 
\and
Radiology, Medical College of Wisconsin, Milwaukee, WI, USA \\
\email{kmkoch@mcw.edu} }

\maketitle  

\begin{abstract}
Quantitative Susceptibility Mapping (QSM) reconstruction is a challenging inverse problem driven by ill conditioning of its field-to -susceptibility transformation. State-of-art QSM reconstruction methods either suffer from image artifacts or long computation times, which limits QSM clinical translation efforts. To overcome these limitations, a non-locally encoder-decoder gated convolutional neural network is trained to infer whole brain susceptibility map, using the local field and brain mask as the inputs. The performance of the proposed method is evaluated relative to synthetic data, a publicly available challenge dataset, and clinical datasets. The proposed approach can outperform existing methods on quantitative metrics and visual assessment of image sharpness and streaking artifacts. The estimated susceptibility maps can preserve conspicuity of fine features and suppress streaking artifacts. The demonstrated methods have potential value in advancing QSM clinical research and aiding in the translation of QSM to clinical operations. 

\keywords{QSM \and Deep learning \and QSM inversion}
\end{abstract}

\section{Introduction}

Quantitative Susceptibility Mapping (QSM) is a MR post-processing technique that estimates underlying tissue magnetic susceptibilities \cite{wang2015quantitative}. QSM has been used to study iron content, blood products, neurodegenerative diseases, brain tumors, and mild traumatic brain injury \cite{haacke2005imaging,zhang2015quantitative,deistung2013quantitative}. Susceptibility maps are generated from MRI data by extracting Larmor frequency shifts from complex MR signals  and solving for the source tissue susceptibility. QSM reconstruction is ill-posed due to the singularity of the dipole kernel that connects susceptibility sources to induced magnetic field offset components in the direction of the main polarizing magnetic field. Current approaches to solve the QSM inverse problem either suffer streaking artifacts, quantification errors, or long computation times, which hinder QSM clinical translation.  

To overcome these limitations, a deep convolutional neural network for QSM inversion is described and analyzed. The presented QSM inversion approach is denoted as QSMInvNet. QSMInvNet approach is evaluated on 100 synthetic datasets, a QSM challenge dataset, and clinical data acquired using a clinically susceptibility weighted imaging (SWI) protocol. The quantitative performance of the neural network is compared with commonly utilized inversion approaches, including Truncated K-Space Division (TKD) inversion~\cite{shmueli2009magnetic},Fast Algorithm for Nonlinear Susceptibility Inversion (FANSI)~\cite{milovic2018fast} and Morphology Enabled Dipole Inversion (MEDI)~\cite{liu2012morphology}. 

\section{Methods}

\subsection{Neural Network Design} 
\subsubsection{Training Data} 

Multiple-Orientation QSM datasets, such as Calculation of Susceptibility through Multiple Orientation Sampling (COSMOS)~\cite{liu2009calculation} or Susceptibility Tensor Imaging (STI)~\cite{liu2010susceptibility} are often treated as QSM golden-standard estimates. However, it is expensive and time-consuming to acquire enough COSMOS or STI data for training of QSM neural networks. Furthermore, COSMOS and STI datasets remain estimates that do not have gold-standard validations (i.e. direct measurements of tissue magnetism). 
In our approach, we utilize one \emph{in-vivo} COSMOS dataset and data augmentation to get large numbers of training data. The COSMOS dataset is from 2016 ISMRM QSM challenge~\cite{langkammer2018quantitative}, which was acquired with a fast 3-dimensional gradient-echo scans, 12 different head orientations, 1.06mm isotropic voxels on a 3T scanner. Elastic transforms are applied to geometrically distort the susceptibility map. In addition, randomly sized and placed geometric shapes, such as ellipsoids, spheres, cuboids and cylinders with random susceptibility values and random orientations are randomly placed on the augmented susceptibility map. Local contrast change was applied to susceptibility map as well for data augmentation as well. The local fields are calculated using the well-defined forward dipole convolution relationship. 5000 synthetic data were simulated using this approach for training.   It is important to note that the COSMOS data is only utilized for a reference susceptibility distribution in this approach.  The precise accuracy of this COSMOS QSM estimate is not of substantial importance, as the network is effectively trained by the well-defined forward dipole-based prediction.  This is a key novelty of the QSM neural network design approach described in this work.      

\begin{figure}[H]
\begin{center}
\vspace{-15pt}
\includegraphics[width=0.5\textwidth]{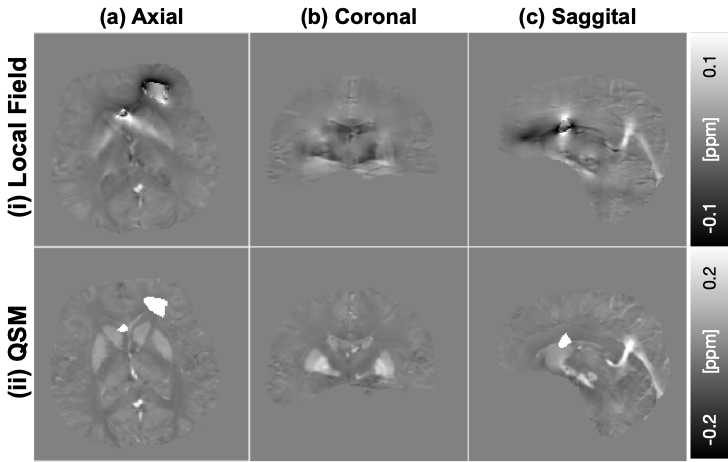}
\caption{Illustration of training data.}
\label{QSMTrainingData}
\vspace{-20pt}
\end{center}
\end{figure}

\subsubsection{Neural Network Architecture and Training} 
A 3D convolutional neural network with encoder-decoder architecture was trained to perform QSM inversion, using whole brain tissue fields and brain masks as the inputs. Gated convolution was utilized in the neural network design, with LeakyReLU as feature activation and Sigmoid for gating values to learn the specific spatial information for susceptibility estimation. Dilated gated convolution was applied to increase the receptive fields. A non-local block was used to enlarge the receptive fields to the entire image and improve the non-local susceptibility estimation of large region with large susceptibility values\cite{wang2018non}. The last layer of the network was a convolutional layer with linear activation to generate the estimated susceptibility output.

\begin{figure}[H]
\begin{center}
\vspace{-15pt}
\includegraphics[width=0.8\textwidth]{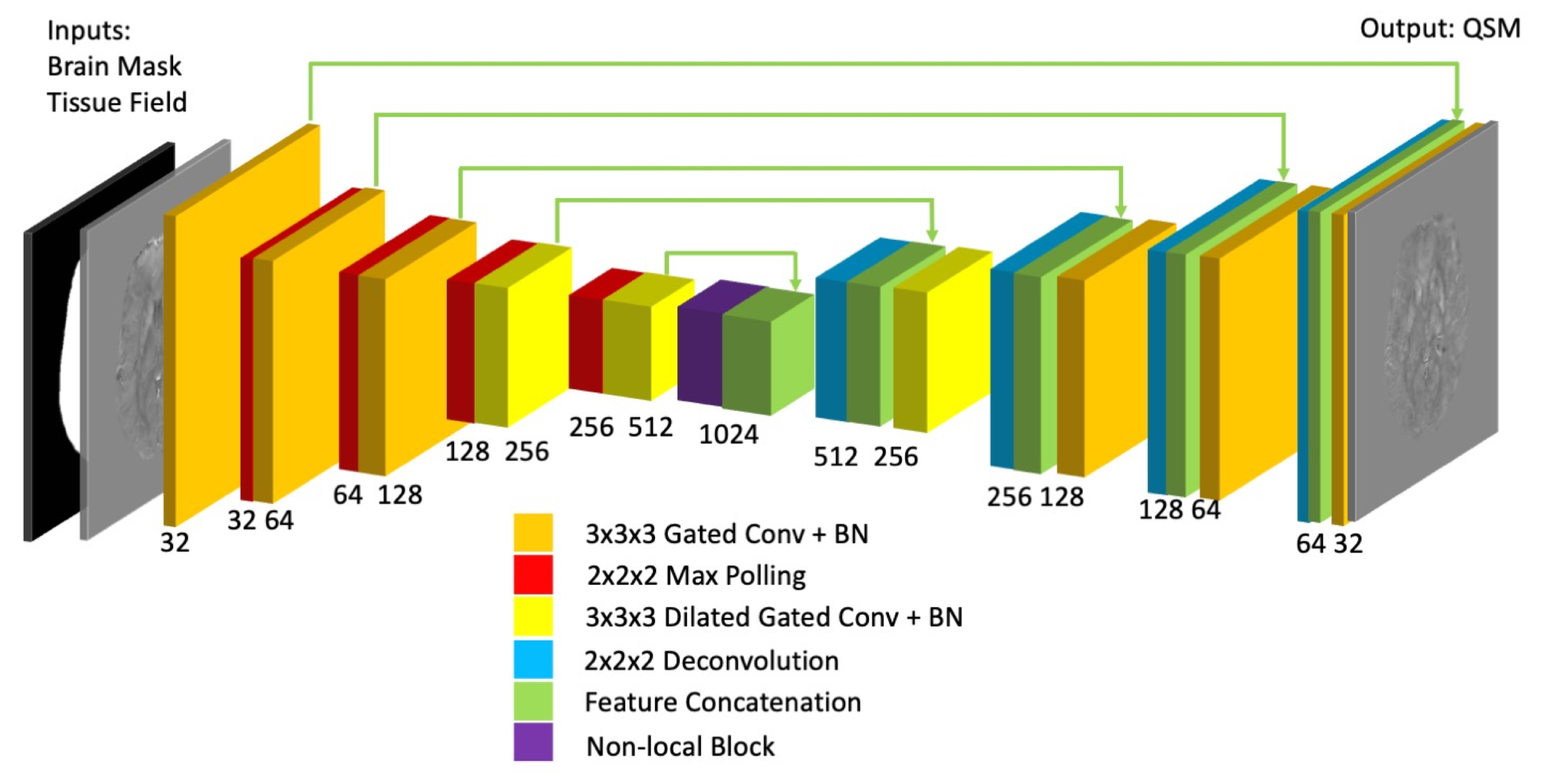}
\caption{Network structure of QSMInvNet. A 3D encoder-decoder architecture was designed with 6 gated convolutional layers (kernel size 3x3x3, dilated rate 1x1x1), 3 gated convolutional layers (kernel size 3x3x3, dilated rate 2x2x2), 4 max pooling layers (pool size 2x2x2, stride size 2x2x2), 4 deconvolutional layers (kernel size 3x3x3, stride size 2x2x2), 1 non-local block, 9 normalization layers, 5 feature concatenations, and 1 convolutional layer (kernel size 3x3x3, linear activation).}
\label{QSMInvNet}
\vspace{-20pt}
\end{center}
\end{figure}

For the synthetic testing and QSM challenge data sets, QSMInvNet input shapes were set to 160x160x160 with voxel size 1.06x1.06x1.06 mm$^3$. For clinical data sets, the input shapes ere set to 256x256x64 with voxel sizes 0.76x0.76x3.0 mm$^3$. L1 loss between the training label and outputs ere utilized as a loss function. The RMSprop optimizer was used for the deep learning training. The initial learning rate was set as 0.0001, with exponential decay at every 200 steps. Two NVIDIA Tesla k40 graphics processing units (GPUs) were used for training with a batch size 2. The neural network was trained and evaluated using Keras with Tensorflow as a backend. 

\subsection{Performance Evaluation} 

\subsubsection{Synthetic Data} 
100 simulated data sets generated in similar fashion to the training data without containing randomly inserted geometric shapes were used to test the performance of QSMInvNet compared with TKD, FANSI, and MEDI. The results were evaluated against the ground-truth susceptibility input using root mean squared error (RMSE), high-frequency error norm (HFEN), and structural similarity (SSIM) index.

\subsubsection{QSM Challenge Dataset} 
Using this gold-standard evaluation dataset, QSMInvNet performance was compared to TKD, FANSI, and MEDI approaches. TKD results were provided publicly by the QSM challenge organizers. All methods were evaluated against the "gold standard" STI (3,3) component computational result provided with the challenge dataset~\cite{liu2010susceptibility}.

\subsubsection{Clinical Data} 
One hundred clinical QSM data were acquired using gradient echo T2 star weighted angiography (SWAN, GE) at a 3T MRI scanner (GE Healthcare MR750) with data acquisition parameters: in-plane data acquisition matrix 288x224, FOV 22cm, slice thickness 3mm, autocalibrated parallel imaging factors 2x1, number of slices 46-54, first echo time 12.6ms, number of echoes 7, echo spacing 4.1ms, flip angle 15\textdegree, TR 39.7ms, total scan time about 2 minutes.  

The SWI images were processed by vendor reconstruction algorithms. The raw k-space data were saved for offline QSM processing. Multi-echo real and imaginary data were reconstructed from k-space data, with reconstruction matrix size 288x288, voxel size 0.76x0.76x3.0 mm$^3$. Larmor offset field maps were obtained by fitting of multi-echo phases. Brain masking was performed using FSL brain extraction tool. After background field removal using regularization enabled sophisticated harmonic artifact reduction for phase data (RESHARP)\cite{sun2014background} with spherical radius 6mm, QSM inversion was performed using TKD, FANSI, MEDI, and QSMInvNet.  

For the purposes of performance evaluation, with TKD, a threshold of 0.20 was manually chosen; for FANSI, ${\mu_1}$ and ${\alpha_1}$ were set to 1e-2, 2e-4 respectively; for MEDI, the regularization factor was set to 1000.

\section{Result}

In Table.\ref{SytheticDataMetic}, QSMInvNet achieved the best score in RMSE, HFEN, and SSIM compared with TKD, FANSI, and MEDI for all 100 of the synthetic data sets.


\begin{table}[!ht]
\centering
\vspace{-10pt}
\caption{\label{SytheticDataMetic}QSM reconstruction quality metrics for synthetic data.}
\vspace{0.0in}
\begin{tabular}{cccccc}
\hline
\multicolumn{1}{|c}{} & \multicolumn{1}{|c}{TKD} & \multicolumn{1}{|c}{FANSI} & \multicolumn{1}{|c}{MEDI} & \multicolumn{1}{|c|}{QSMInvNet} \\
\hline 
\multicolumn{1}{|c}{{RMSE} ($\%$)} & \multicolumn{1}{|c}{33.1$\pm$0.29} & \multicolumn{1}{|c}{39.0$\pm$1.41} & \multicolumn{1}{|c}{34.7$\pm$0.37} & \multicolumn{1}{|c|}{\bf{22.5$\pm$0.42}} \\
\hline
\multicolumn{1}{|c}{{HFEN} ($\%$)} & \multicolumn{1}{|c}{35.2$\pm$0.30} & \multicolumn{1}{|c}{36.7$\pm$1.6} & \multicolumn{1}{|c}{38.1$\pm$0.26} & \multicolumn{1}{|c|}{\bf{25.0$\pm$0.58}} \\
\hline 
\multicolumn{1}{|c}{{SSIM} (0-1)} & \multicolumn{1}{|c}{0.967$\pm$0.002} & \multicolumn{1}{|c}{0.951$\pm$0.004} & \multicolumn{1}{|c}{0.933$\pm$0.004} & \multicolumn{1}{|c|}{\bf{0.978$\pm$0.001}} \\
\hline \\
\end{tabular}
\begin{flushleft}
\end{flushleft}
\vspace{-40pt}
\end{table}

\begin{figure}[H]
\begin{center}
\vspace{-10pt}
\includegraphics[width=0.9\textwidth]{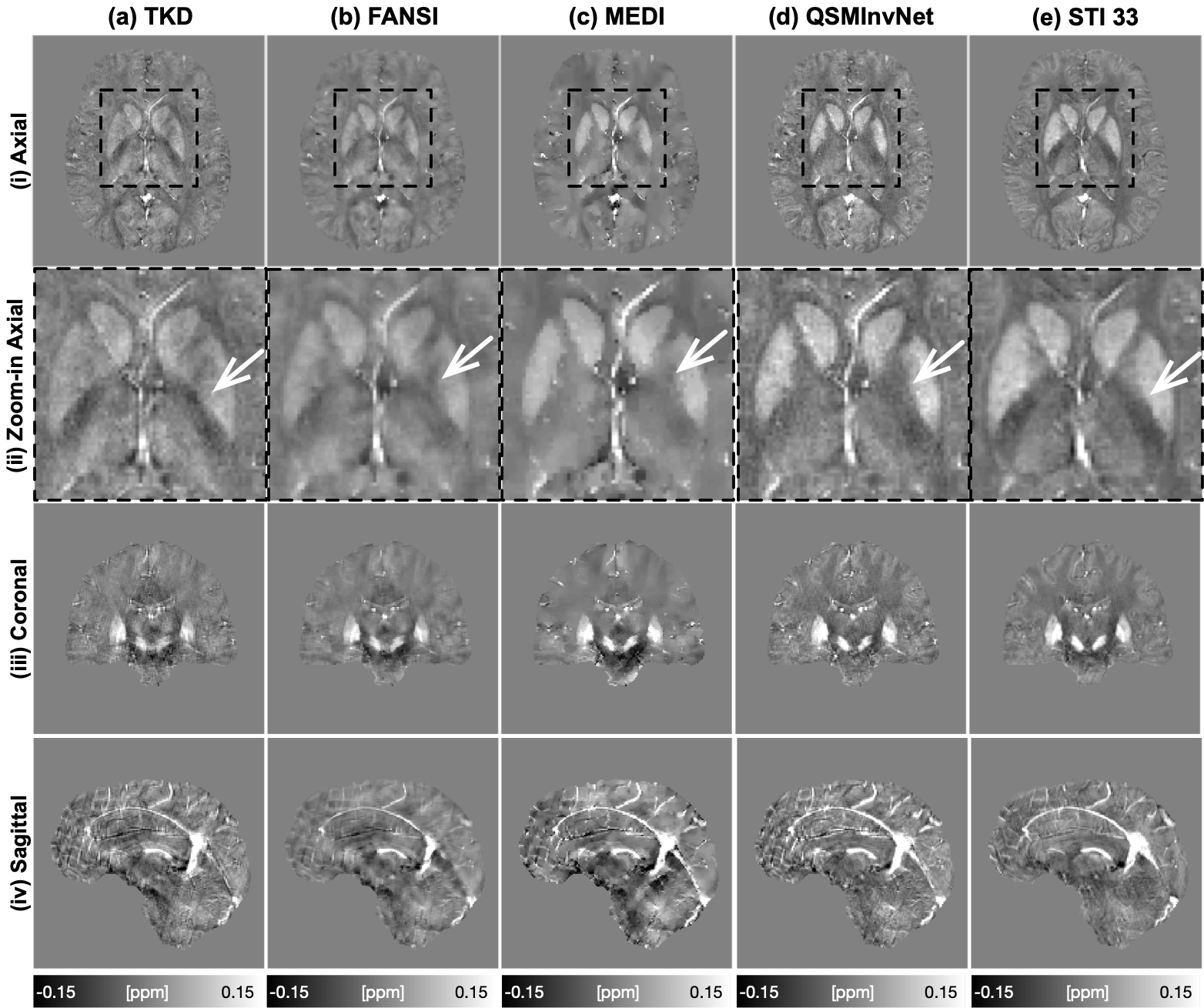}
\caption{QSM maps of the QSM challenge dataset reconstructed by the four methods.}
\label{qsmchannlge} 
\vspace{-20pt}
\end{center}
\end{figure}

In Fig.\ref{qsmchannlge}, QSM maps of the QSM challenge dataset reconstructed by TKD (a), FANSI (b), MEDI (c), and QSMInvNet (d) are compared with the ground truth (e). From the zoom-in axial images (ii), substantial image blurring and conspicuity loss of fine details is clearly visible in TKD, FANSI, and MEDI images. QSMInvNet maps have superior image sharpness and well-preserved details, as indicated by white arrows. 

\begin{figure}[H]
\begin{center}
\vspace{-20pt}
\includegraphics[width=0.9\textwidth]{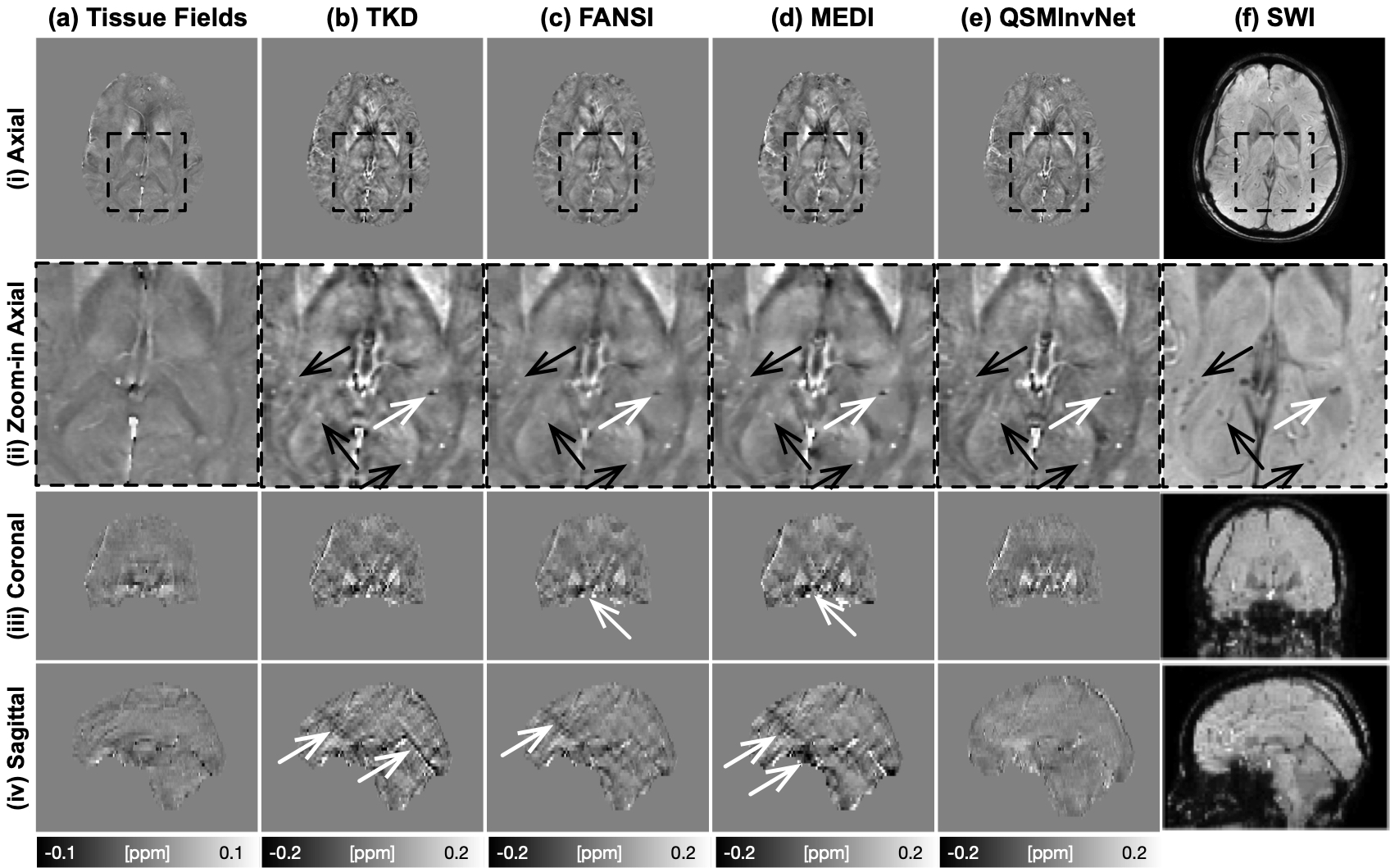}
\caption{ Total fields (a), tissue fields (b), QSM images (c-f), SWI images (g) from a 34-year-old subject with subdural fluid collection and history of meningioma resection.}
\label{srcMircobleeds} 
\vspace{-30pt}
\end{center}
\end{figure}

In Fig.\ref{srcMircobleeds}, QSM and SWI images of a patient with subdural fluid collection are illustrated. In zoom-in axial (ii), a few hypointense regions of SWI image (black arrow) is hyperintense in QSM, indicating it iron deposition or hemorrhage. One small calcification (white arrows) is hypointense on SWI image and diamagnetic on QSM image. From zoom-in axial (ii), QSMInvNet images show best image sharpness. From the coronal and saggital view (iii, iv), TKD, FANSI, and MEDI show streaking artifacts (white arrows).

\begin{figure}[H]
\begin{center}
\vspace{-20pt}
\includegraphics[width=\textwidth]{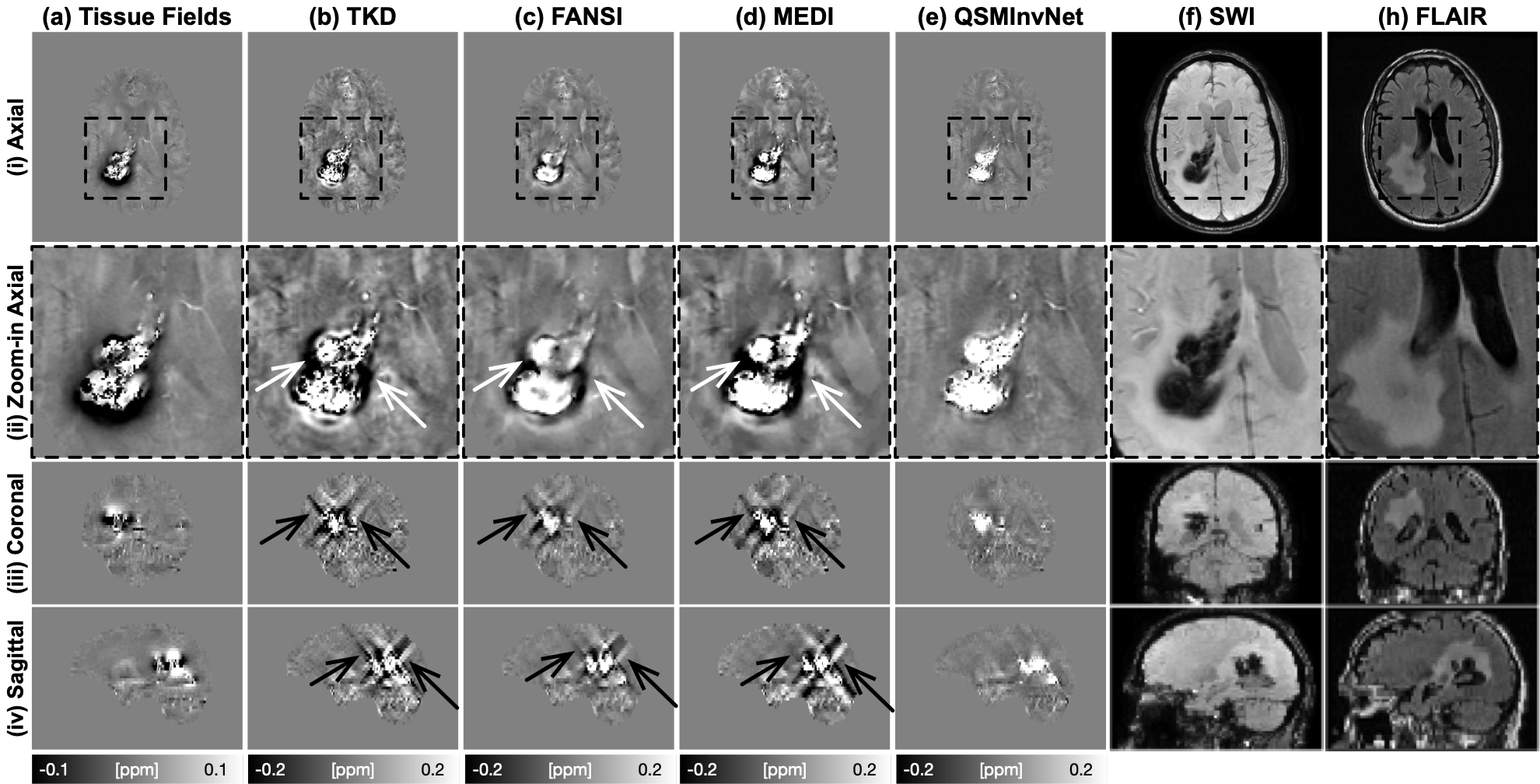}
\caption{ Total fields (a), tissue fields (b), QSM images (c-f), SWI images (g), FLAIR images (h) from a  54-year-old  subject  with  poststereotactic radiosurgery (SRS) brain metastasis.}
\label{srcHemorrahge} 
\vspace{-20pt}
\end{center}
\end{figure}

\begin{figure}[H]
\begin{center}
\vspace{-20pt}
\includegraphics[width=\textwidth]{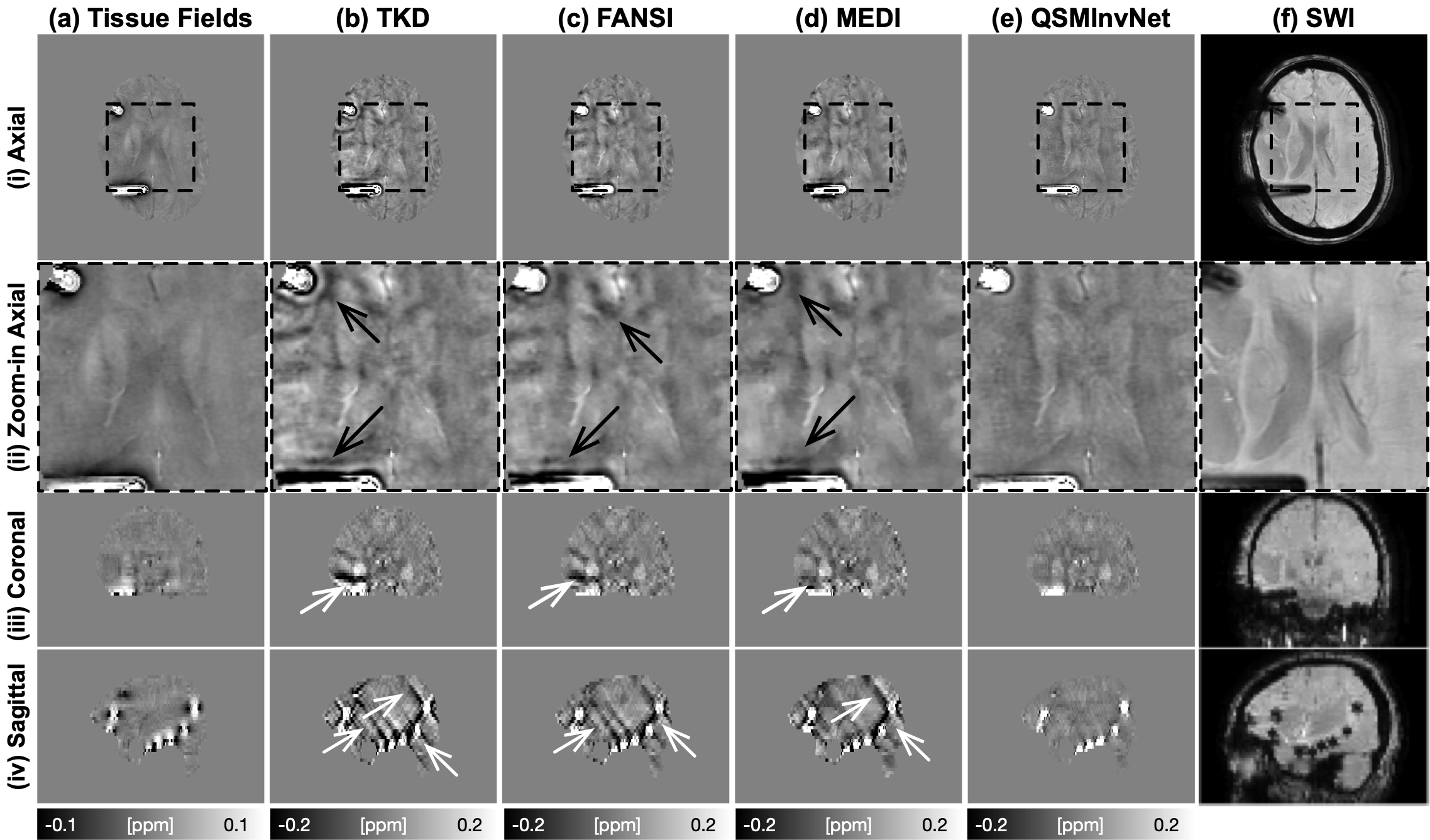}
\caption{ Total fields (a), tissue fields (b), QSM images (c-f), SWI images (g), FLAIR images (h) from a  37-year-old subject  with surgical planning.}
\label{srcSurgical} 
\vspace{-20pt}
\end{center}
\end{figure}

In Fig.\ref{srcHemorrahge}, the SWI and QSM images of a patient with SRS brain metastasis are illustrated. It is clearly visible black shading artifacts in axial plane and severe streaking artifacts in sagittal plane in TKD, FANSI, and MEDI images. QSMInvNet images show the best image quality with high image sharpness and no streaking artifacts. From the Fig.\ref{srcSurgical}, QSMInvNet images show no shading or streaking artifacts, while TKD, FANSI, and MEDI suffers from image blurring, shading artifacts, and streaking artifacts.

\section{Discussion}

In this work, QSMInvNet, a neural network for QSM reconstruction, was described and evaluated on synthetic data, public challenge data, and routine clinical data. For synthetic data sets with a gold standard reference, the proposed method achieved better quantitative performance than TKD, FANSI, and MEDI on RMSE, HFEN, and SSIM. The public challenge and clinical data sets results showed that QSMInvNet can produce high quality susceptibility maps with superior image sharpness and no-visible streaking artifacts. Clinical examples demonstrate that QSMInvNet can utilize raw SWI data from existing standard of care exams to reconstruct high-quality QSM images that preserve the fine details and suppress streaking artifacts. Compared with current QSM reconstruction methods, QSMInvNet requires no regularization parameter tuning for QSM inversion. It can perform QSM reconstruction in real-time on GPU hardware, which can help facilitate the use of QSM clinical practice. In addition, QSMInvNet images can preserve fine structures and suppress streaking artifacts.


The presented QSMInvNet approach introduces several important innovations. First, it performs whole brain high-resolution QSM inversion using a neural network. Compared with patch-based neural networks, it avoids patch merging and tiling artifacts. Second, it utilizes a non-local block to increase the receptive fields and capture long-range information for non-local susceptibility estimation. Third, it uses gated convolutions to learn spatial information that help in performing inner brain and brain-boundary susceptibility estimation. 

This feasibility study has also demonstrated the ability to use existing standard of care  SWI raw data to reconstruct QSM for clinical utility. This offers the possibility of QSM use in clinical operation without any additional scans beyond current standard of care protocols. Combining SWI magnitude and QSM estimation images may offer new diagnostic capabilities to assist radiological interpretation. In particular, it is well-known that SWI suffers from blooming artifacts and difficulties in differentiating  calcifications and hemosiderin. QSM can overcome these limitations of SWI, which can expand the roles of SWI and QSM in neuroradiology clinical and research arenas. In the Fig.\ref{srcMircobleeds}, the calcification is easily differentiated in QSM maps. From the Fig.\ref{srcHemorrahge}, QSMInvNet results show no shading artifacts or streaking artifacts around the lesions, while also  preserving the details of fine structures.

\section{Conclusion}
In summary, a deep QSM inversion approach has been demonstrated. It can substantially improve brain susceptibility estimation. This capability opens up a wide array of QSM investigations using clinically acquired SWI data to derive and analyze QSM maps across a host of clinical neurological conditions. 

\bibliographystyle{unsrt}
\bibliography{references}

\begin{thebibliography}{10}
\providecommand{\url}[1]{\texttt{#1}}
\providecommand{\urlprefix}{URL }
\providecommand{\doi}[1]{https://doi.org/#1}

\bibitem{deistung2013quantitative}
Deistung, A., Schweser, F., Wiestler, B., Abello, M., Roethke, M., Sahm, F.,
  Wick, W., Nagel, A.M., Heiland, S., Schlemmer, H.P., et~al.: Quantitative
  susceptibility mapping differentiates between blood depositions and
  calcifications in patients with glioblastoma. PLoS ONE  \textbf{8}(3),
  e57924 (2013)

\bibitem{haacke2005imaging}
Haacke, E.M., Cheng, N.Y., House, M.J., Liu, Q., Neelavalli, J., Ogg, R.J.,
  Khan, A., Ayaz, M., Kirsch, W., Obenaus, A.: Imaging iron stores in the brain
  using magnetic resonance imaging. Magnetic Resonance Imaging  \textbf{23}(1),
   1--25 (2005)

\bibitem{langkammer2018quantitative}
Langkammer, C., Schweser, F., Shmueli, K., Kames, C., Li, X., Guo, L., Milovic,
  C., Kim, J., Wei, H., Bredies, K., et~al.: Quantitative susceptibility
  mapping: report from the 2016 reconstruction challenge. Magnetic Resonance in
  Medicine  \textbf{79}(3),  1661--1673 (2018)

\bibitem{liu2010susceptibility}
Liu, C.: Susceptibility tensor imaging. Magnetic Resonance in Medicine
  \textbf{63}(6),  1471--1477 (2010)

\bibitem{liu2012morphology}
Liu, J., Liu, T., de~Rochefort, L., Ledoux, J., Khalidov, I., Chen, W.,
  Tsiouris, A.J., Wisnieff, C., Spincemaille, P., Prince, M.R., et~al.:
  Morphology enabled dipole inversion for quantitative susceptibility mapping
  using structural consistency between the magnitude image and the
  susceptibility map. NeuroImage  \textbf{59}(3),  2560--2568 (2012)

\bibitem{liu2009calculation}
Liu, T., Spincemaille, P., De~Rochefort, L., Kressler, B., Wang, Y.:
  Calculation of susceptibility through multiple orientation sampling (cosmos):
  a method for conditioning the inverse problem from measured magnetic field
  map to susceptibility source image in mri. Magnetic Resonance in Medicine
  \textbf{61}(1),  196--204 (2009)

\bibitem{milovic2018fast}
Milovic, C., Bilgic, B., Zhao, B., Acosta-Cabronero, J., Tejos, C.: Fast
  nonlinear susceptibility inversion with variational regularization. Magnetic
  resonance in medicine  \textbf{80}(2),  814--821 (2018)

\bibitem{shmueli2009magnetic}
Shmueli, K., de~Zwart, J.A., van Gelderen, P., Li, T.Q., Dodd, S.J., Duyn,
  J.H.: Magnetic susceptibility mapping of brain tissue in vivo using mri phase
  data. Magnetic Resonance in Medicine  \textbf{62}(6),  1510--1522 (2009)

\bibitem{sun2014background}
Sun, H., Wilman, A.H.: Background field removal using spherical mean value
  filtering and tikhonov regularization. Magnetic Resonance in Medicine
  \textbf{71}(3),  1151--1157 (2014)

\bibitem{wang2018non}
Wang, X., Girshick, R., Gupta, A., He, K.: Non-local neural networks. In:
  Proceedings of the IEEE Conference on Computer Vision and Pattern
  Recognition. pp. 7794--7803 (2018)

\bibitem{wang2015quantitative}
Wang, Y., Liu, T.: Quantitative susceptibility mapping (qsm): decoding mri data
  for a tissue magnetic biomarker. Magnetic Resonance in Medicine
  \textbf{73}(1),  82--101 (2015)

\bibitem{zhang2015quantitative}
Zhang, J., Liu, T., Gupta, A., Spincemaille, P., Nguyen, T.D., Wang, Y.:
  Quantitative mapping of cerebral metabolic rate of oxygen (cmro2) using
  quantitative susceptibility mapping (qsm). Magnetic Resonance in Medicine
  \textbf{74}(4),  945--952 (2015)

\end{thebibliography}

\newpage
\section*{Supplement}

\begin{figure}[H]
\begin{center}
\vspace{-10pt}
\includegraphics[width=\textwidth]{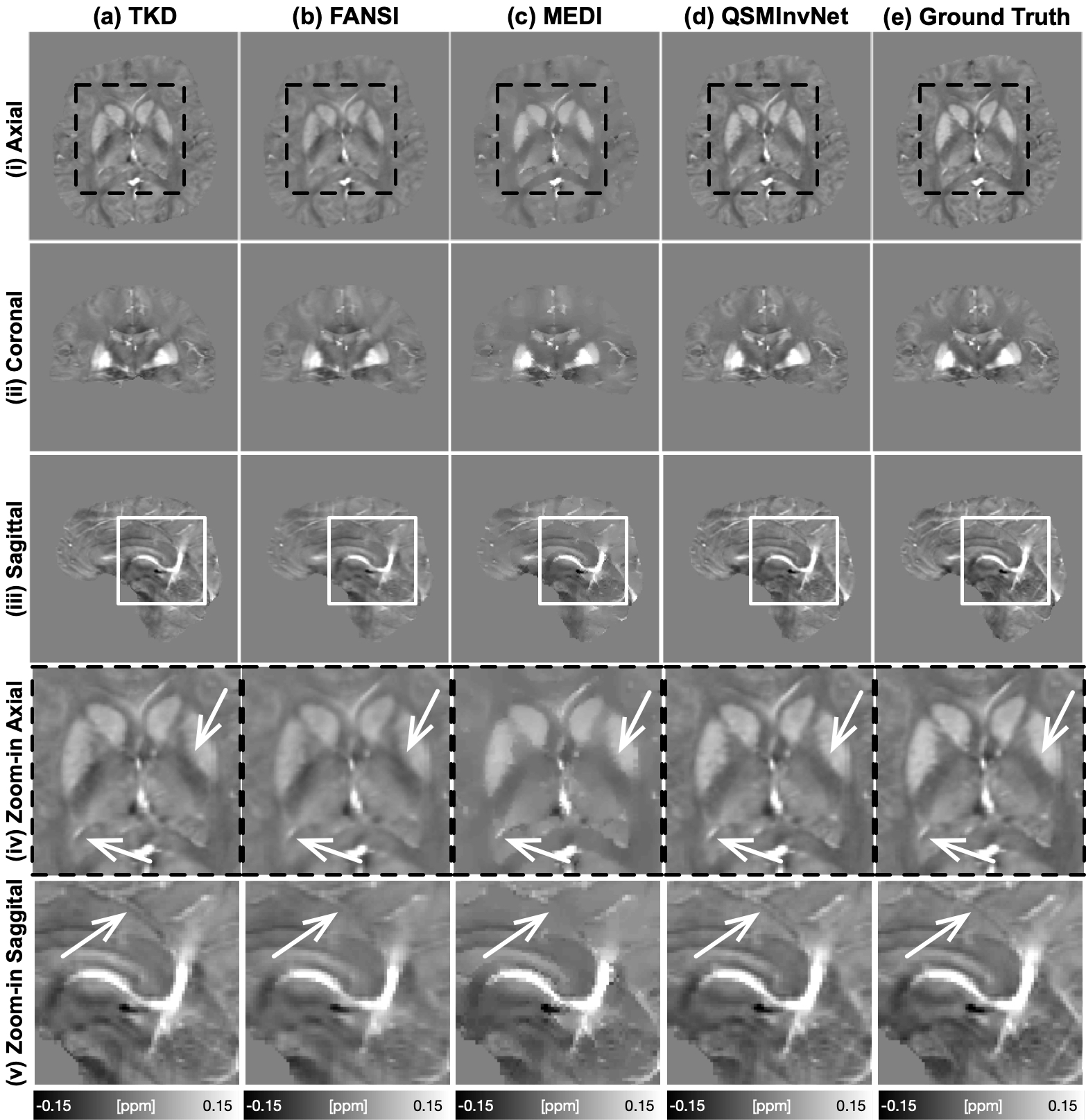}
\caption{QSM maps of one synthetic testing data. From the zoom-in axial and saggital planes (iv-v), it is clearly visible that TKD, FANSI, and MEDI suffer from image blurring. QSMInvNet can well preserve the tissue boundary and fine details. }
\label{testdata_qsm} 
\vspace{-20pt}
\end{center}
\end{figure}

\begin{figure}[H]
\begin{center}
\vspace{-10pt}
\includegraphics[width=\textwidth]{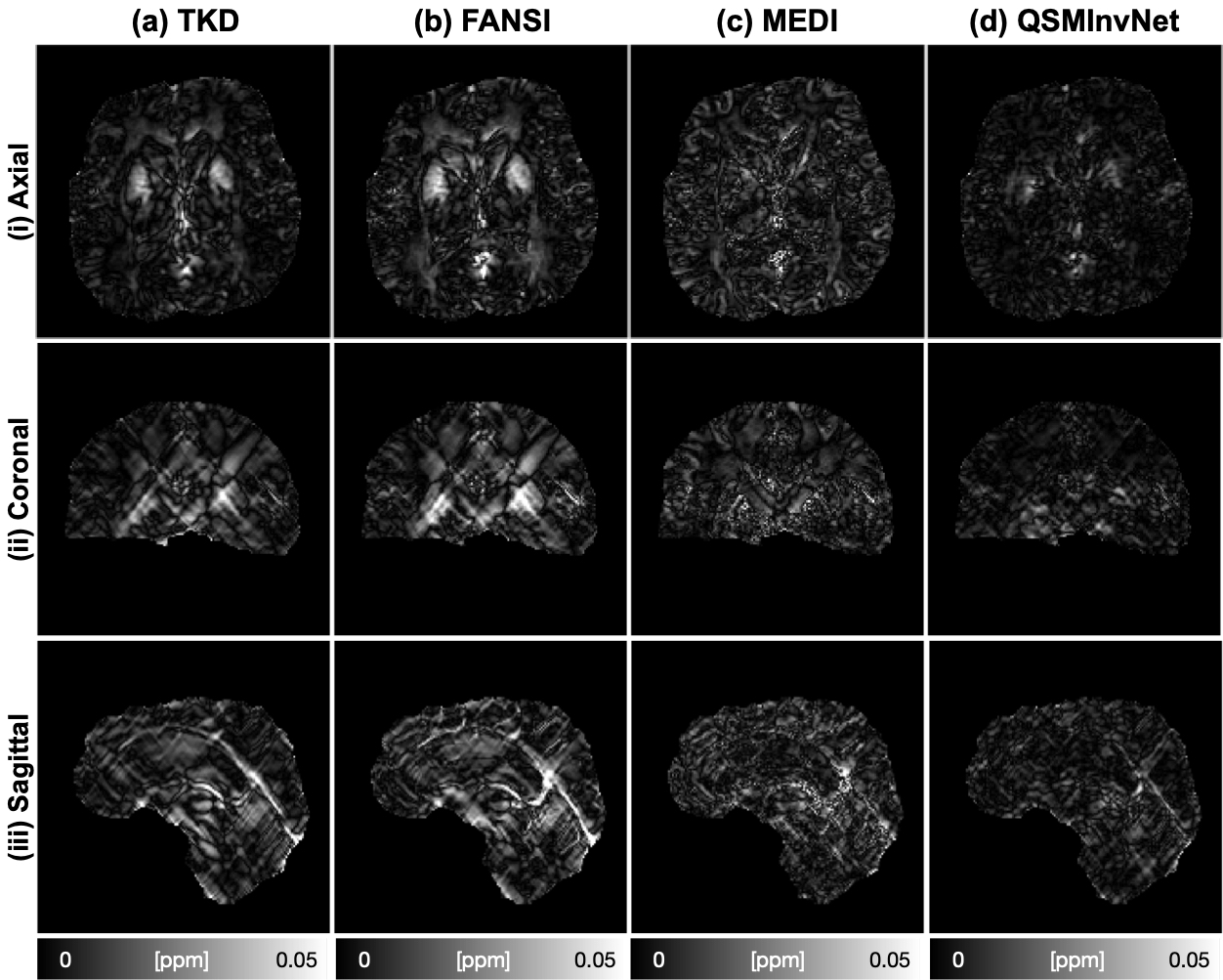}
\caption{The residual error maps of Fig.\ref{testdata_qsm}. QSMInvNet results have the least errors compared with TKD, FANSI, and MEDI results. QSMInvNet has the least residual error compared with other methods.}
\label{testdata_errmap} 
\vspace{-20pt}
\end{center}
\end{figure}


\begin{figure}[H]
\begin{center}
\vspace{-10pt}
\includegraphics[width=\textwidth]{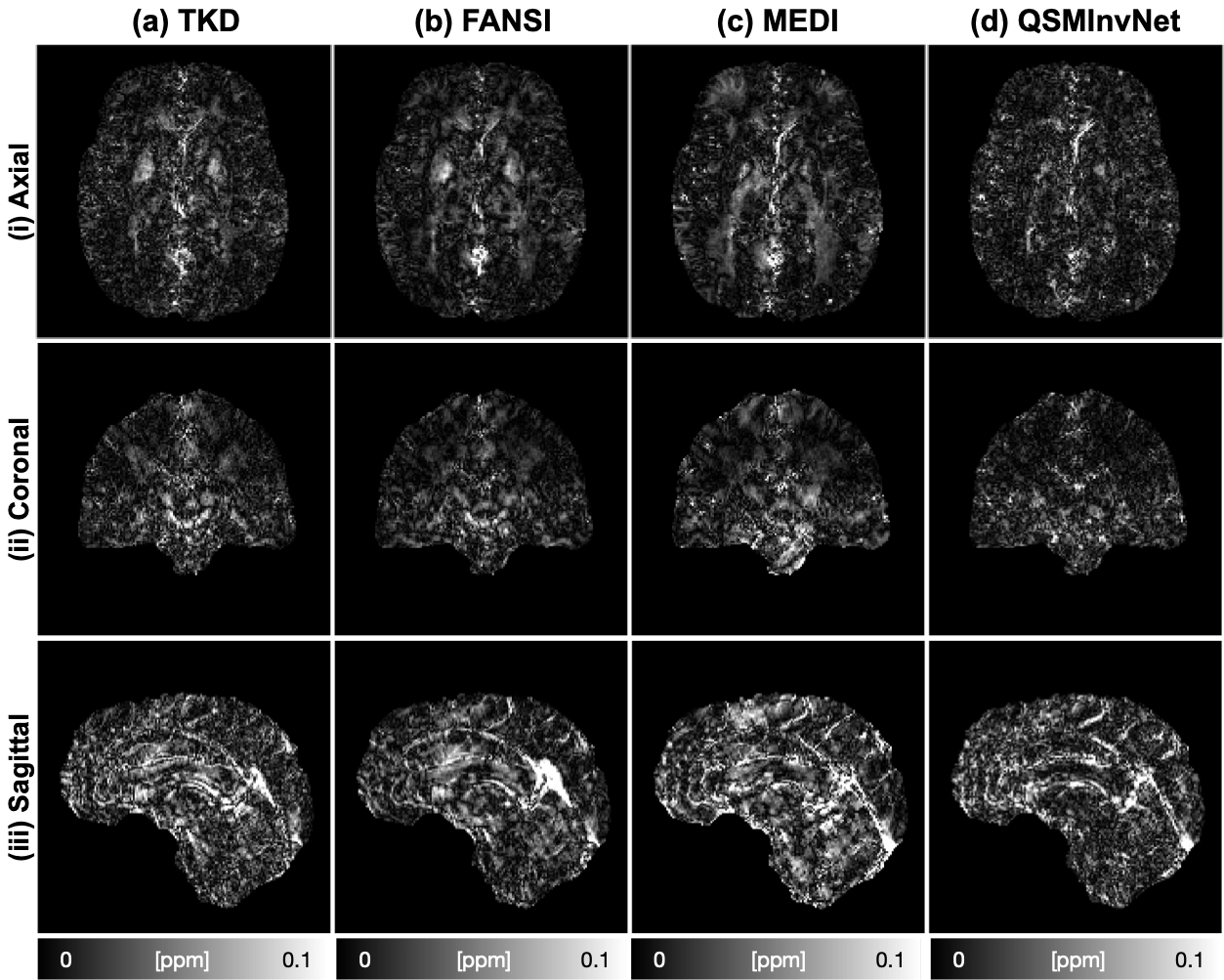}
\caption{The residual error maps of QSM challenge datasets with comparision with STI 33. It is obviously shown that QSMInvNet have the least residual errors.}
\label{qsmchannlge_errmap} 
\vspace{-20pt}
\end{center}
\end{figure}


\begin{figure}[H]
\begin{center}
\vspace{-10pt}
\includegraphics[width=0.9\textwidth]{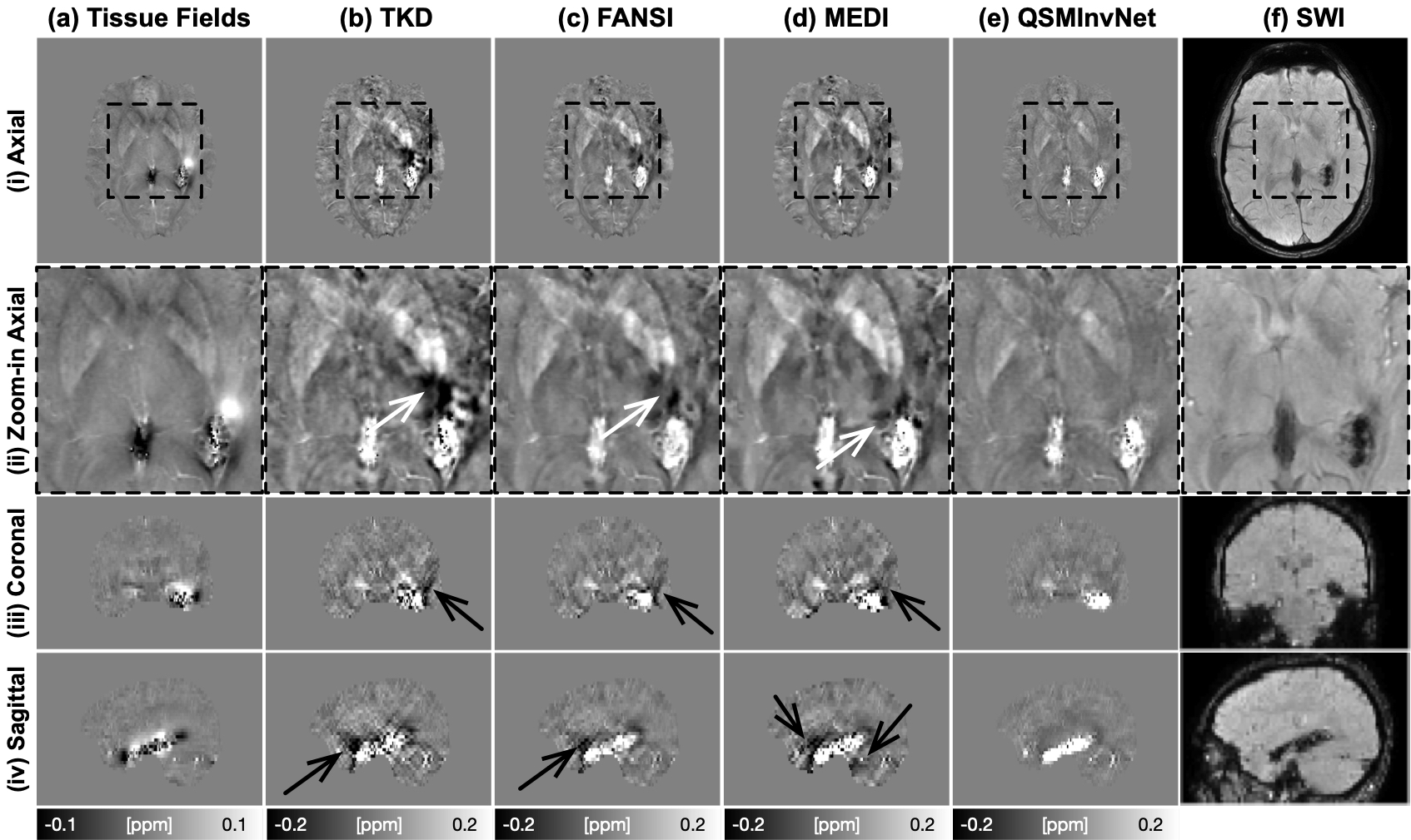}
\caption{ Tissue fields (a), QSM images (b-e), SWI images (f) from a 28-year-old subject with left mesial temporal lesion and Neurofibromatosis Type-1. In the zoom-in axial (ii), TDK, FANSI, and MEDI images have shading artifacts close to the bleeding region (white arrows). QSMInvNet images show non-visible artifacts around the bleeding regions. In the coronal and saggital planes (iii, iv), shading artifacts and streaking artifacts is clearly visible in TKD, FANSI, and MEDI results (black arrows). }
\label{p1388} 
\vspace{-20pt}
\end{center}
\end{figure}

\begin{figure}[H]
\begin{center}
\vspace{-10pt}
\includegraphics[width=0.9\textwidth]{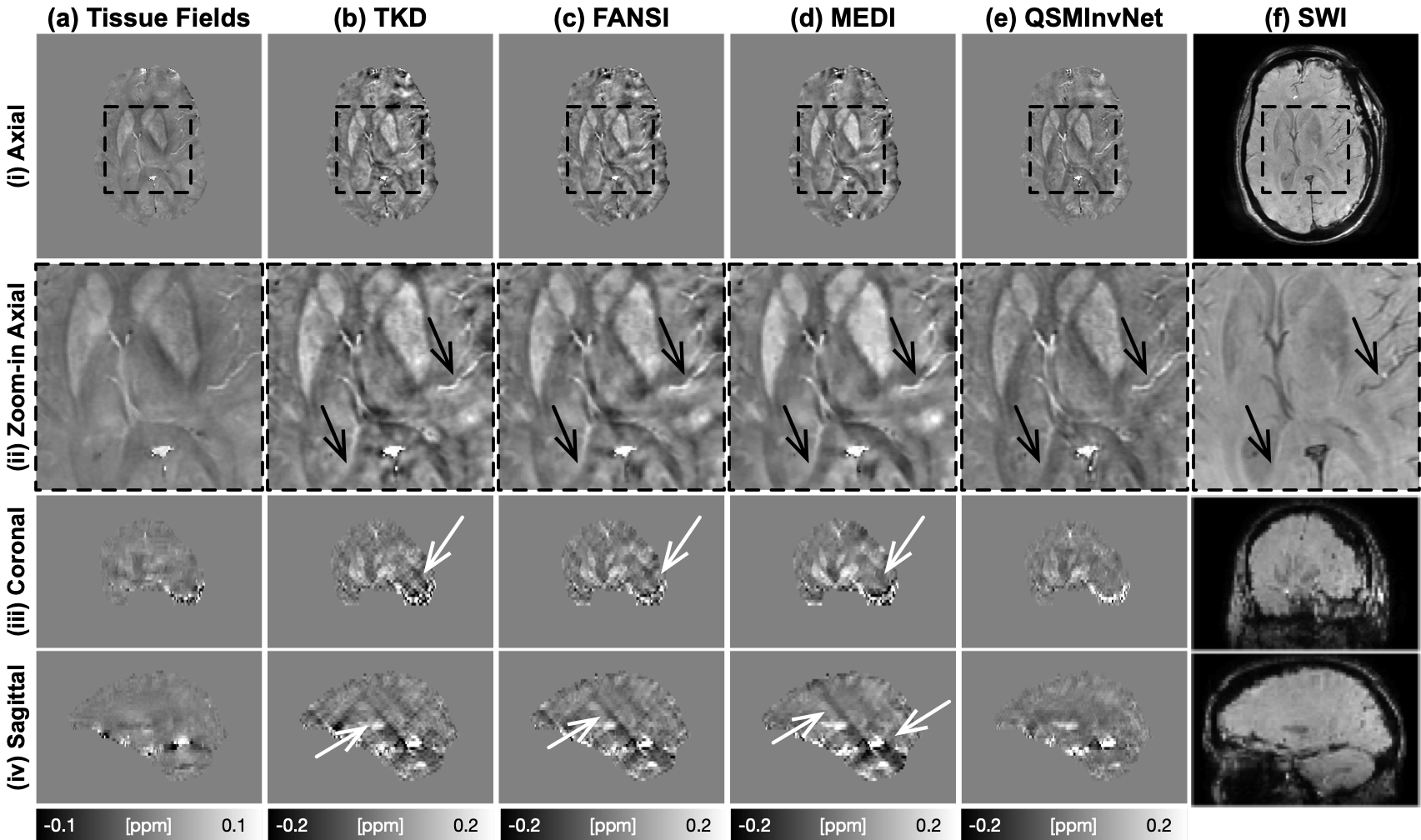}
\caption{ Tissue fields (a), QSM images (b-e), SWI images (f) from a 56-year-old subject with hemorrhagic intracranial metastases. In zoom-in axial, brain vessels is clearly visible in QSM images. Image blurring is clearly visible in TKD, FANSI, and MEDI images, while QSMInvNet show better image sharpness (black arrows). In the coronal and saggital planes (iii, iv), streaking artifacts is clearly visible (white arrows). QSMInvNet results show no shading and streaking artifacts. }
\label{p1619} 
\vspace{-20pt}
\end{center}
\end{figure}



\begin{figure}[H]
\begin{center}
\vspace{-10pt}
\includegraphics[width=0.9\textwidth]{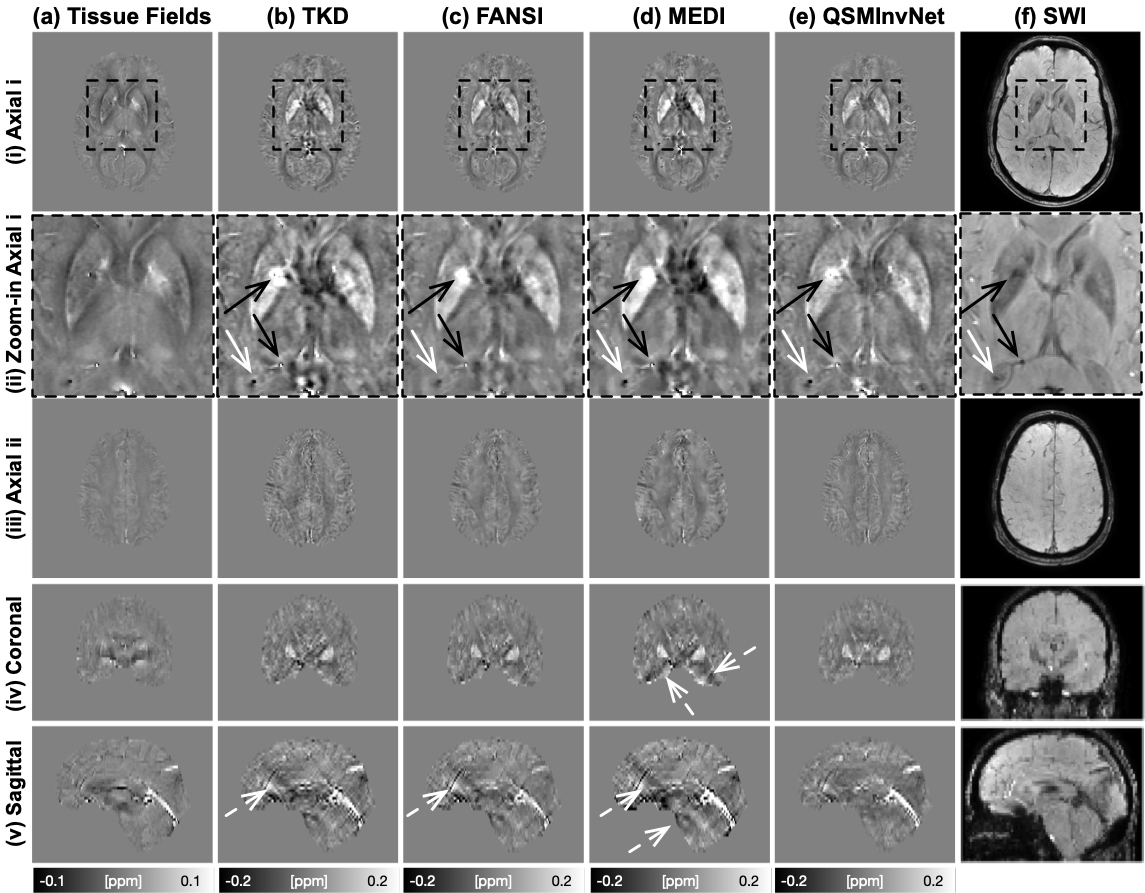}
\caption{ Total fields (a), tissue fields (b), QSM images (c-f), SWI images (g) from a 58-year-old subject with lung cancer. Two microbleeds/iron deposition and one calcifications all appear as black hypointense regions in SWI images, making it difficult to differentiate one from another. In QSM images, microbleeds/iron deposition (paramagnetic) show as bright/hyperintense regions, while calcifications (diamagnetic) are dark/hypointense regions. Compared with TKD, FANSI, and MEDI, QSMInvNet can produce QSM images with super image sharpness and no streaking artifacts, as shown in (ii, iv).}
\label{p1701_resharp} 
\vspace{-20pt}
\end{center}
\end{figure}



\begin{figure}[H]
\begin{center}
\vspace{-10pt}
\includegraphics[width=0.9\textwidth]{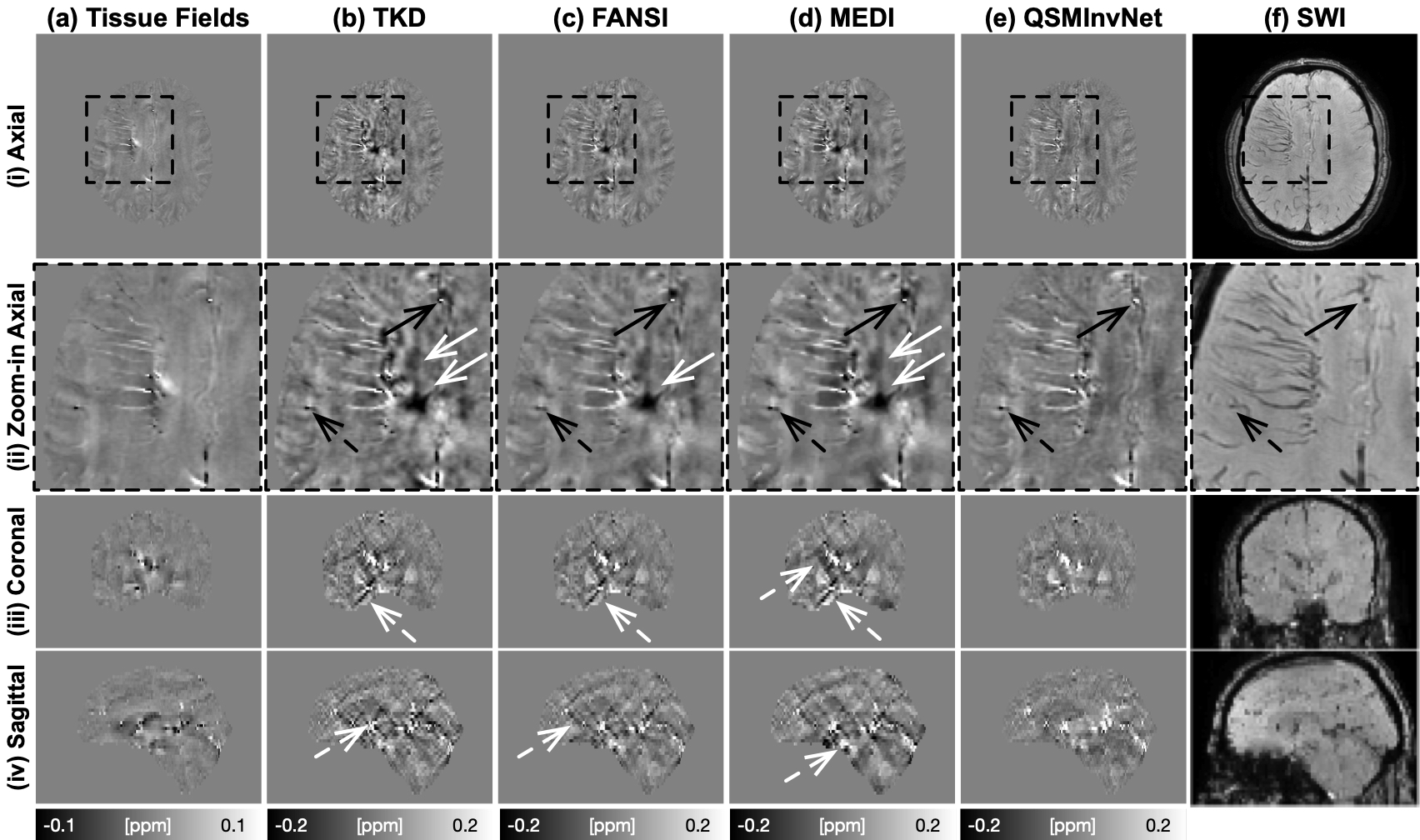}
\caption{ Tissue fields (a), QSM images (b-e), SWI images (f) from a 34-year-old patient with cerebrovas-cular accident (CVA) and spinal cord etiology of leftupper extremity and left lower extremity weakness. In zoom-in axial (ii), TKD, FANSI, and MEDI images show bright in QSM images and dark/hypointense in SWI image (dark solids arrows). A small calcification is dark/hypointense on SWI image and diamagnetic on QSM images (black dash arrows). In saggital plane (iv), streaking artifacts are clearly visible in TKD and MEDI images (white dash arrows).}
\label{p618} 
\vspace{-20pt}
\end{center}
\end{figure}

\begin{figure}[H]
\begin{center}
\vspace{-20pt}
\includegraphics[width=0.9\textwidth]{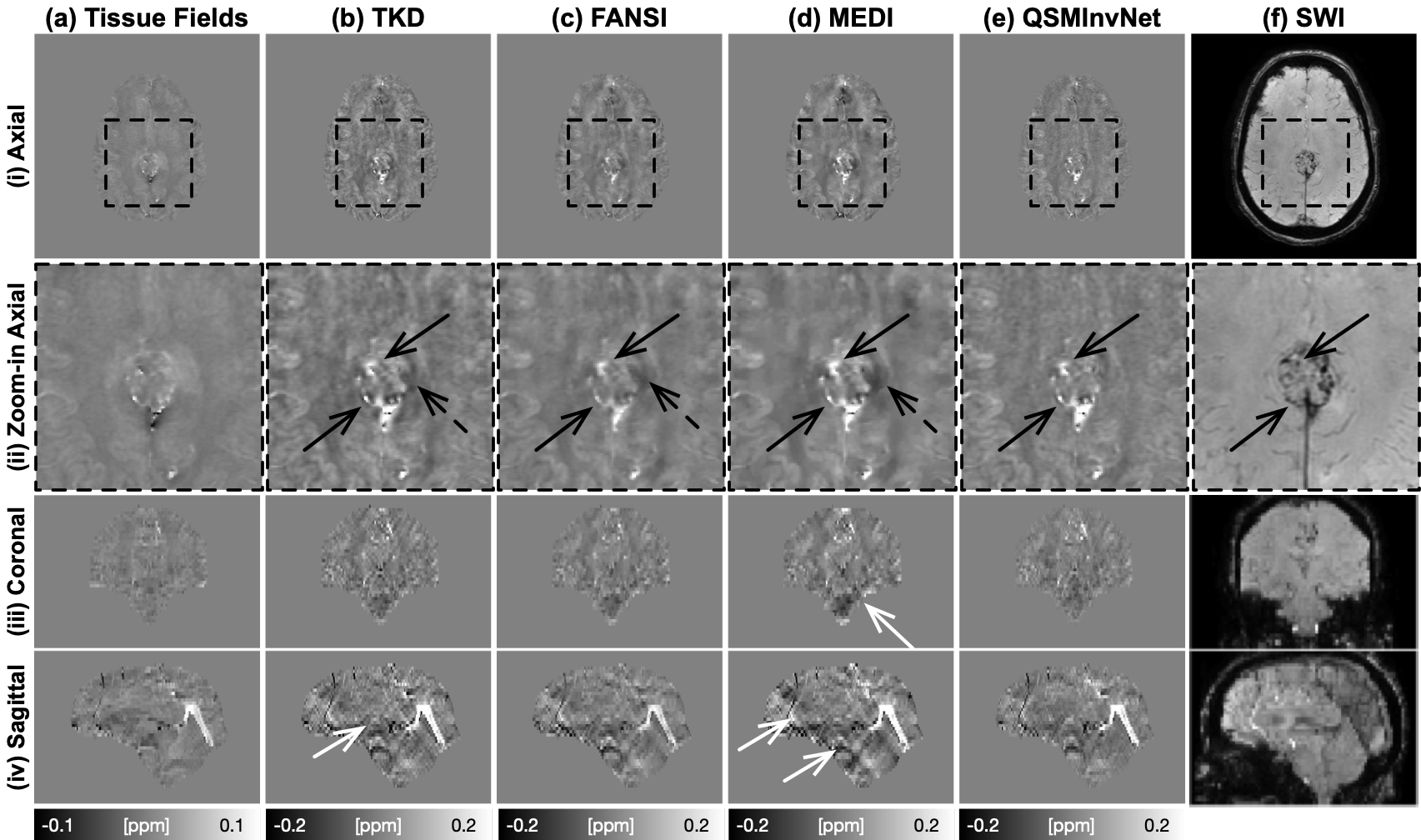}
\caption{ Tissue fields (a), QSM images (b-e), SWI images (f) from a 59-year-old subject with brain tumor. The brain tumor region of SWI image is hyperintense in QSM, indicating hemorrhage in the brain tumor. In zoom-in axial (ii), TKD, FANSI, and MEDI images show black shading artifacts (black dash arrows) around the brain tumor. Compared with other methods, QSMInvNet show better image sharpness (black arrows). In saggital plane (iv), streaking artifacts are clearly visible in TKD and MEDI images (white arrows).}
\label{p618} 
\vspace{-20pt}
\end{center}
\end{figure}

\end{document}